\documentclass{article}

\usepackage[final,nonatbib]{nips_2016}

\usepackage{amssymb}
\usepackage{graphicx}
\usepackage{todonotes}
\usepackage{amsmath} 
\usepackage{algpseudocode}
\usepackage{capt-of}
\usepackage{subfig} 

\usepackage{url}

\DeclareMathAlphabet{\mathpzc}{OT1}{pzc}{m}{it}
\newcommand{\argmax}{\operatornamewithlimits{arg\ max}} 

\newcommand{\perfGain}[1]{\operatorname{perfGain}\big( #1 \big)}
\newcommand{\perf}[1]{\operatorname{perf}\big( #1 \big)}

\newcommand{\expExpPerf}[1]{\operatorname{expPerf}\big( #1 \big)}

\newcommand{\E}[2]{\operatornamewithlimits{\mathbb{E}}_{#1} \left[ #2 \right]}

\title{Probabilistic Active Learning for \\Active Class Selection}

\author{
  Daniel Kottke \\
  Intelligent Embedded Systems \\
  University Kassel, Germany\\
  \texttt{daniel.kottke@uni-kassel.de} \\
  \And
  Georg Krempl \\
  Knowledge Management \& Discovery \\
  University Magdeburg, Germany \\
  \texttt{georg.krempl@ovgu.de}
  \And 
  Marianne Stecklina \quad Cornelius Styp von Rekowski \quad Tim Sabsch \quad Tuan Pham Minh\\
  University Magdeburg, Germany \\
  \texttt{\{marianne.stecklina, cornelius.styp, tim.sabsch, tuan.pham\}@ovgu.de}
  \And
  \And 
  Matthias Deliano \\
  Leibniz Institute for\\
  Neurobiology, Magdeburg \\
  \texttt{deliano@lin-magdeburg.de} 
  \And 
  Myra Spiliopoulou \\
  KMD Group  \\
  University Magdeburg \\
  \texttt{myra@cs.uni-magdeburg.de}
  \And
  Bernhard Sick\\
  IES Group\\
  University of Kassel\\
  \texttt{bsick@uni-kassel.de} \\
}

\begin{document}
\maketitle

\begin{abstract}

In machine learning, active class selection (ACS) algorithms aim to actively select a class and ask the oracle to provide an instance for that class to optimize a classifier's performance while minimizing the number of requests. 
In this paper, we propose a new algorithm (PAL-ACS) that transforms the ACS problem into an active learning task by introducing pseudo instances. These are used to estimate the usefulness of an upcoming instance for each class using the performance gain model from probabilistic active learning.
Our experimental evaluation (on synthetic and real data) shows the advantages of our algorithm compared to state-of-the-art algorithms. It effectively prefers the sampling of difficult classes and thereby improves the classification performance.
\end{abstract}

\section{Introduction} \label{intro}


Many classification systems found in application areas such as economy, medical research or neurobiology require human labeling effort during training. 
As this is time-consuming and expensive, the field of active learning (AL) emerged \cite{Settles2012}. 
Here, the aim is to actively choose only the most informative instances from a large pool of unlabeled data and successively request their label. 
As a result, good classification performance is reached with less training instances compared to passively feeding arbitrary instances to the classifier.

In this article, we address a related field called active class selection (ACS) \cite{Lomasky2009}.
Instead of selecting an unlabeled instance and acquiring its label, ACS methods request a yet unseen instance by selecting its class. 
On the one hand, the degree of freedom is much smaller in ACS compared to AL as there are normally less classes than instances. 
On the other hand, less information is available to decide what is beneficial for training, e.g. unlabeled instances to approximate the data distribution are missing.

\begin{figure}[h]
  \centering
  \includegraphics[width=.8\textwidth]{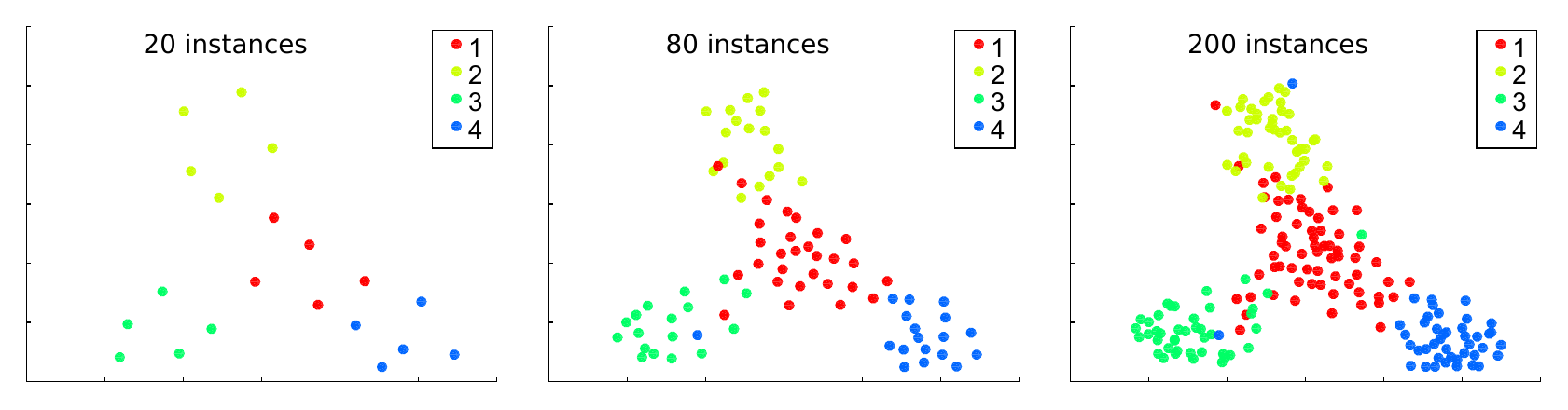}
  \caption{t-SNE plot for different learning stages in an ACS sampling process}
  \label{fig:acs_process}
\end{figure}

Why is ACS a topic worth researching? 
A vivid example for the application of ACS is the training of brain computer interfaces for motoric prostheses. 
To train such a prosthesis, an impaired patient has to imagine motoric movements \cite{hohne} for example of his fingers while his brain activity is recorded. 
Fig.~\ref{fig:acs_process} shows different learning stages of such an exemplary ACS process. 
In the beginning, the algorithm only knows the number of classes (fingers). 
If certain classes (fingers) are hard to distinguish in the data (here class 1), learning should focus on these classes. 
By requesting the patient to generate more training instances of these classes instead of spending time on already learned classes, a good classification performance is achieved earlier -- an achievement that enables the patient to perform otherwise impossible tasks \cite{hohne}.

As visualized by this example, ACS is useful whenever classes vary in difficulty. 
We contribute a new method to the field of ACS that is able to identify the difficulty of classes. 
The core idea is to transform the ACS task into an active learning problem which enables the applicability of well-understood active learning paradigms:
In each step, we simulate the generation of instances (called pseudo instances). 
Next, we take the performance gain function used in probabilistic active learning~\cite{KottkeKrempl2016ECAI} to determine the expected usefulness of a new instance request for every class. 
Finally, we request an instance from the class with highest usefulness. 
This is repeated until a maximal number of instances (budget) is reached.

The rest of the article is structured as follows. In Sec.~\ref{rel}, we discuss the literature on active class selection, followed by our new method PAL-ACS. We provide a pseudo code and discuss properties of the approach using an example. After an evaluation on multiple datasets in Sec.~\ref{eval}, we finally conclude our work.

\section{Related Work} \label{rel}
Active classification systems have the ability to request relevant information from external sources. With respect to the type of requested information, different approaches are distinguished \cite{AttenbergMelvilleProvostEtAl2011}. The most intensively researched ones actively select instances for labeling from an oracle. The aim of these so-called active learning methods is to select those instances whose labels will improve the classification performance the most \cite{Settles2012}. Scope of this paper is the inverse setting of
active learning, which is called active class selection (ACS) \cite{Settles2010}: Here, the active component is able to select a class from which subsequently an unknown instance (feature vector) is generated. 

The idea of ACS is to distribute the number of instances per class such that a certain level of classification performance is reached with the lowest number of requested instances \cite[p.~29]{AttenbergMelvilleProvostEtAl2011}. The work presented in  \cite{LomaskyBrodleyAerneckeEtAl2007} (see also \cite{LomaskyBrodleyAerneckeEtAl2006}) mentions different techniques to determine this class distribution for acquisition chunks. First, Lomasky et al.~\cite{LomaskyBrodleyAerneckeEtAl2007} propose to use a \emph{uniform} distribution and the \emph{Original Proportion} (that usually is not known) as baselines. Furthermore, they perform what they call $f$-fold cross validation on the already seen chunks to use the results for the next chunk: The approach \emph{Inverse} distributes the information according to the inverse of the class accuracy. An extension of this is called \emph{Accuracy Improvement}. It distributes the values according to the accuracy difference between the two most current chunks. The \emph{Redistricting} method counts the number of labels that have been flipped (these instances are marked as redistricted) by adding the most recent chunk to the training set. Here, the upcoming instances are distributed with respect to the number of redistricted instances of the true classes. 

Wu and Parsons \cite{WuParsons2011} applied the previous algorithms \emph{Inverse} and \emph{Accuracy Improvement} to arousal classification. Later, they extended this article in \cite{WuLanceParsons2013} and improved the approach \emph{Inverse} to be applicable for incremental stream acquisitions along with the addition of a constraint that two consecutive new training examples are from different classes. In her PhD thesis \cite{Lomasky2009}, Lomasky extended her work by two more methods: \emph{Risk} estimates the sensitivity of error that is induced by adding new instances of a certain class, and \emph{Sensitivity} measures the stability of class decisions. As these methods are only mentioned in the PhD thesis yielding mediocre results, we solely consider the former ones in our evaluation.

As mentioned in the introduction, our new method transforms the ACS task into an active learning problem \cite{Settles2012} and uses probabilistic active learning \cite{KremplKottkeLemaire2015OPAL}, which is assigned to the group of decision theoretic methods.
Classical decision theoretic approaches simulate the acquisition of every possible label and evaluate their effect on the classification error using an evaluation set \cite{RoyMccallum2001}. In \cite{Chapelle2005}, Chapelle observed that these error reduction estimates have issues with unreliable posterior estimates at the beginning. Thus, he suggests using a beta-prior to shift posterior values with less labeled information towards equal posterior probabilities. 
Probabilistic active learning \cite{KremplKottkeLemaire2015OPAL} reduces the computational complexity of the previous methods by using local statistics (number of nearby labels) to estimate the usefulness of a labeling candidate. This approach models the true posterior probability with a Beta distribution in order to include the reliability of the posterior. Probabilistic active learning has been extended for multi-class problems in \cite{KottkeKrempl2016ECAI} and is discussed in Sec.~\ref{sec:method:palacs} in more detail.

\section{Our Method} 
\label{pacs}

In this section, we propose our new method called \emph{Probabilistic Active Learning for Active Class Selection (PAL-ACS)}. 
The first subsection gives a detailed description of our algorithm including the necessary background on probabilistic active learning. To support the understanding of the algorithm, we show a visualization of its behavior and provide a pseudo code which can be used for implementing the approach in the second and third subsection.




\subsection{Probabilistic Active Learning for Active Class Selection (PAL-ACS)}
\label{sec:method:palacs}
The main idea of our algorithm is to estimate the gain in classification performance for each class $y \in Y = \{1, \dots , C\}$ when requesting one additional instance of that class $y$. 
Then, we request an instance $x^*$ of the best class $y^*$ and add this new training sample to the training set $\mathcal{L} \gets \mathcal{L} \cup (x^*,y^*)$. 

To estimate the expected gain in performance that a label request would probably induce, probabilistic active learning \cite{KottkeKrempl2016ECAI} provides an effective tool. 
Its performance gain function can be calculated at any location in the feature space, regardless of the fact if there is a real unlabeled instance at this location.
It only requires local statistics, which typically are labeling counts $\vec{k} = (k_1, \dots, k_C)$. A common strategy to determine this vector are kernel frequencies \cite{KottkeKrempl2016ECAI}. These sum up the similarities from the requested location ($x$) to every instance from class $y$. \begin{align}\label{eq:kfe}
  k_i = \mathrm{KFE}(x,\mathcal{L}_i) &= \sum_{\{(x',y') \in \mathcal{L}_i\}} \mathrm{sim}(x,x')
  = \sum_{(x',y') \in \mathcal{L}_i} \mathrm{exp}\left( - \frac{|| x-x' ||^2}{2\sigma^2} \right)\\
  \mathcal{L}_i &= \{(x,y) \in \mathcal{L} \colon y = i\}
\end{align}

In probabilistic active learning \cite{KremplKottkeLemaire2015OPAL}, it is assumed that the similarity function defines a neighborhood around the requested location $x$ which separates the data into being inside resp. outside the neighborhood. Hence, a newly added label of class $y$ would increase $k_y$ by $1$, resp. $n$ new labels increase the corresponding elements in the frequency vector $\vec{k}$ by $n$.
Accordingly, the labeling vector $\vec{l} = (l_1, \dots, l_C)$, $l_i \in \mathbb{N}$, $\sum l_i = m$ is defined such that the cells contain the number of labels that might be added to that neighborhood \cite{KottkeKrempl2016ECAI}. Considering to add $m=2$ labels to a neighborhood for example (let $C=3$), the labeling vectors could be $\vec{l} \in \{(2,0,0),(0,2,0), (0,0,2), (1,1,0), (1,0,1), (0,1,1)\}$.

The performance gain function~(Eq.~\ref{eq:perfGain}) \cite{KottkeKrempl2016ECAI} for label statistics $\vec{k}$ is defined by subtracting the current expected performance ($0$ labels added) from the future expected performance ($m$ labels added). This function includes the addition of \emph{multiple} hypothetical labels ($m>1$). As we only acquire labels successively (one-by-one), we divide the gain by the number of labels $m$ which could be interpreted as the average gain in performance. In our application, the parameter $M$ which sets the upper bound for the so-called local budget has been set to $3$ to avoid high computational time.
\begin{align} \label{eq:perfGain}
  &\perfGain{\vec{k},M} =  \max_{m \leq M} \left( \frac{1}{m} \big(\expExpPerf{\vec{k},m}  - \expExpPerf{\vec{k},0} \big) \right)
\end{align}

The expected performance in Eq.~\ref{eq:expExpGain} is a decision theoretic formulation calculating the expectation values over all possible posterior probabilities $\vec{p}$ and over all possible labeling vectors $\vec{l}$ \cite{KottkeKrempl2016ECAI}. The probability of a posterior probability $P(\vec{p} \mid \vec{k})$ to be true given the current label statistics is derived using the likelihood of the multinomial distribution. The probability of a labeling vector $P(\vec{l} \mid \vec{p})$ to be true is directly given by the multinomial distribution. We optimize the performance in terms of accuracy as given in Eq.~\ref{eq:perf}. More details can be found in \cite{KottkeKrempl2016ECAI}.
\begin{align} \label{eq:expExpGain}
  \expExpPerf{\vec{k},m} &= \E{\vec{p}}{   \E{\vec{l}}{   \perf{\vec{k}+\vec{l} \mid \vec{p}}   }} \\
  \label{eq:perf}
  \perf{\vec{k}+\vec{l} \mid \vec{p}} &= \vec{p}_{\hat{y}} \qquad \quad
  \hat{y} = \argmax(\vec{k}+\vec{l})
\end{align}

In contrast to active learning, we do not have access to a pool of unlabeled instances in ACS.
Thus, we propose to generate pseudo instances $x_p$ to transform the active class selection problem into an active learning task. We then use the pseudo instances to determine the most beneficial class $y^*$ which is selected according to Eq.~\ref{eq:pal-acs:2}. Distributing the pseudo instance randomly or equidistant over the whole feature space, we have to add two weights: 
(1) Similarly to probabilistic active learning, we incorporate an instance's impact on the overall classification performance, i.e. a density weight~\cite{KremplKottkeSpiliopoulou2014DS} ($P(x_p \mid \mathcal{L})$). 
(2) We weight the instance by the probability given that it is assigned to the requested class $y$ to distinguish the classes ($P(x_p \mid \mathcal{L}, y)$). In practice, we use a Monte Carlo approach instead of equidistant sampling as described in Sec.~\ref{pacs:pseudeocode}.
\begin{align} \label{eq:pal-acs:2}
  y^* &= \argmax_y \left( \sum_{x_p} P(x_p \mid \mathcal{L}) \cdot P(x_p \mid \mathcal{L}_y) \cdot \mathrm{perfGain}(KFE(x_p,(\mathcal{L})_i))   \right)
\end{align}

\subsection{Characteristics of PAL-ACS and Example}
\begin{figure}[h]
  \centering
  \includegraphics[width=.45\textwidth]{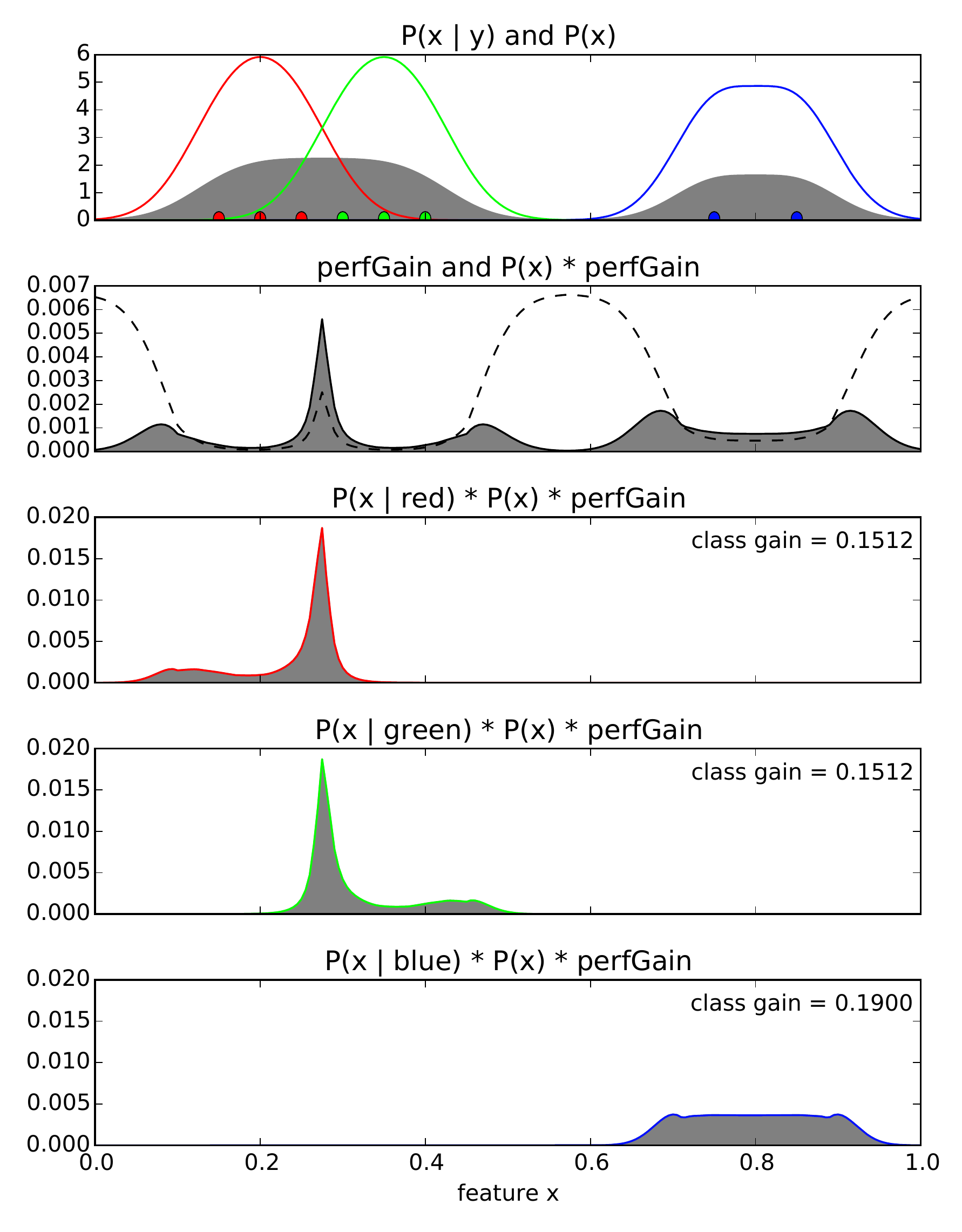}
  \includegraphics[width=.45\textwidth]{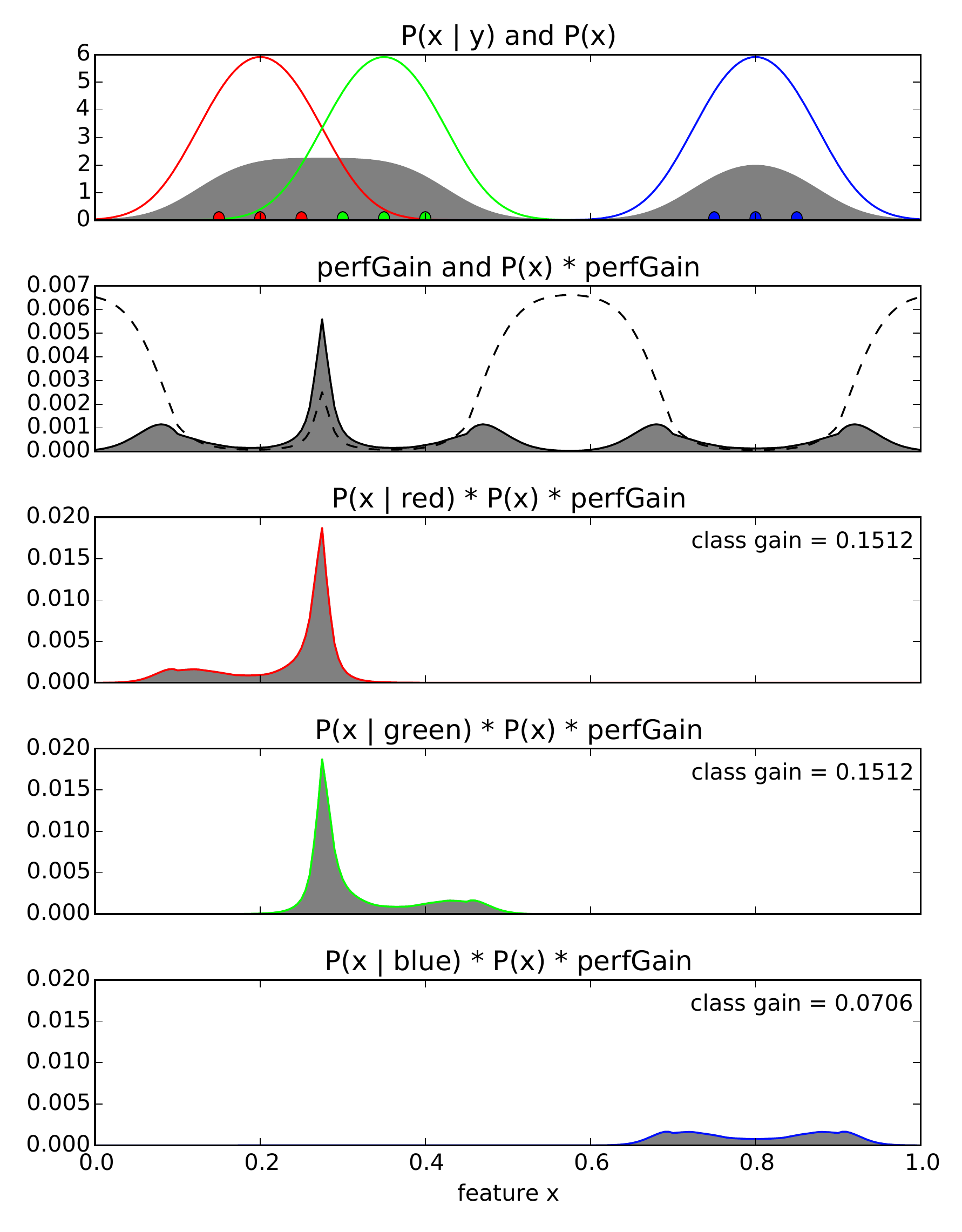}\\ \vspace{2mm}
  \includegraphics[width=.8\textwidth]{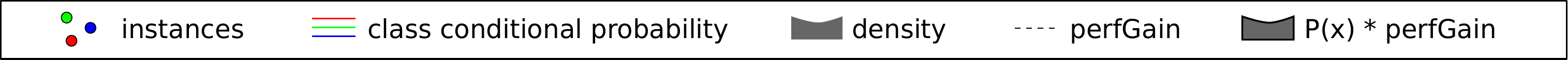}
  \caption{Visualization of \textit{PAL-ACS} for different situations.}
  \label{fig:algo-visu}
\end{figure}

We now discuss PAL-ACS's approach in two exemplary active class selection situations shown in Fig.~\ref{fig:algo-visu}. Both situations are based on a three-class-classification task with a one-dimensional feature space. One class (blue) is well separated from the other two classes (red and green) and can therefore be considered to be easy. Due to an overlap of the other two classes, finding the best decision boundary between them is more difficult. 

The situations shown in the left and right columns are from consecutive selection steps. On the left, 8 instances (3 red, 3 green, 2 blue) have already been acquired, and on the right, there is one additional blue instance.
The upper plots show the location of the instances (colored dots on the x-axis) with its corresponding class (red, green, and blue).
Furthermore, they show the class conditional distributions in the corresponding color and the density as a gray area. The plots in the second horizontal row show the $\mathrm{perfGain}$ function over the whole feature space as a black dashed line, and the density weighted $\mathrm{perfGain}$ as a solid line with gray area. The lower three plots show the density weighted $\mathrm{perfGain}$ (solid black curve from above) additionally weighted with the corresponding class conditional probabilities which build the final score from Eq.~\ref{eq:pal-acs:2}. The numbers in the upper right corners represent the sum of the corresponding values. The class with the maximal value is chosen for the next instance generation.

The difference between both snapshots is the smaller number of blue instances on the left. Thus the uncertainty is higher which is underlined by an higher performance gain value. This lack of information is responsible to have an instance of the blue class requested. On the right, all classes are equally well represented (by 3 instances each). Here, the complexity of the decision boundary and the uncertainty in that specific region is responsible to prefer the red, resp. green class.
As discussed in \cite{KottkeKrempl2016ECAI}, the $\mathrm{perfGain}$ function balances exploration and exploitation by using the number of nearby labels. This also works when using the model for ACS tasks.

\subsection{Implementation and Pseudo Code}
\label{pacs:pseudeocode}

In Fig.~\ref{fig:pseudocode}, we provide the pseudo code of our approach, starting with the sampling of pseudo instances~$\mathcal{X}^p$ in line~\ref{alg:sampling}. Especially for high-dimensional data, an equidistant sampling of the whole feature space exceeds computational capabilities. 
Hence, we use a Monte-Carlo approach in our implementation. From each class, we sample $n_{p} = 25$ pseudo instances from the corresponding density distribution (line~\ref{alg:sampling}). 
The distribution to sample from is determined by a kernel density estimation similar to the frequency estimate's kernel. 
In ACS, it is generally assumed that each class is similarly important (albeit not all are necessarily equally difficult). Therefore, we sample the same number of pseudo instances from each class. 

\begin{figure}[h]
  \begin{algorithmic}[1]
    \State $n_{p} \gets 25$, \quad $M \gets 3$, \quad $\mathcal{L} \gets \{\}$ \Comment{Set initial standard values} \label{fig:algo:startinit}
    \vspace{.5em}
    \While{instance acquisitions left}
      \State $\mathcal{X}^{p} \gets \textrm{SampleFromDensity}(\mathcal{L}, n_p)$ \label{alg:sampling} \Comment{Sample pseudo instances (PP)}
      \For{$i \in \{1,\dots,|\mathcal{X}^{p}|\}$} \label{alg:perfGain:1} \Comment{Calculate pseudo instance's $\textrm{perfGain}$s}
        \State $k_{i,.} \gets \textrm{getKVector}(x^p_i, \mathcal{L})$
        \State $pg_i \gets \textrm{perfGain}(\vec{k}_i,M) / |\mathcal{X}^{p}|$ \Comment{Multiply density weight}
      \EndFor \label{alg:perfGain:2}
      \vspace{.5em}
      \For{$y \in \{1,\dots,C\}$} \label{alg:classcond:1} \Comment{Summarize weighted $\textrm{perfGain}$s} 
        \State $g_y \gets 0$
        \For{$i \in \{1,\dots,|\mathcal{X}^{p}|\}$} 
          \State $g_y \gets g_y + pg_i \cdot k_{i,y} / (\sum_{j=1}^{|\mathcal{X}^{p}|} k_{j,y}) $
        \EndFor
      \EndFor \label{alg:classcond:2}
      \vspace{.5em}
      \State $y^* \gets \argmax_{y} (g_y)$ \label{alg:argmax} \Comment{Select optimal class}
      \State $x^* \gets \textrm{requestInstance}(y^*)$
      \State $\mathcal{L} \gets \mathcal{L} \cup (x^*,y^*)$ \label{alg:appendinst}
    \EndWhile
  \end{algorithmic}
  \caption{Pseudo code of the PAL-ACS algorithm.}
  \label{fig:pseudocode}
\end{figure}

In the for-loop (lines~\ref{alg:perfGain:1}-\ref{alg:perfGain:2}), we estimate the kernel frequency vector as defined in Eq.~\ref{eq:kfe} and calculate the corresponding performance gain (see Eq.~\ref{eq:perfGain}) for each pseudo instance. As all values are generated from the data, each pseudo instance is now equally probable. As we sampled the instance according to the density, the density weight is a simple division by the number of pseudo points. 
In lines~\ref{alg:classcond:1}-\ref{alg:classcond:2}, we weight this density-weighted performance gain with the class conditional probability and sum all values for each class separately. Finally, we select the best class gain $g_y$ and request a corresponding instance (lines~\ref{alg:argmax}-\ref{alg:appendinst}).

The parameter $M$ of the $\mathrm{perfGain}$ function is a parameter of the probabilistic approach that defines the so-called local budget. In case $M$ additional labels are not able to change the classification decision in a neighborhood, the $\mathrm{perfGain}$ is zero. This might lead to inconsistencies in the learning process. In our experiments, a value of $M=3$ was sufficient (higher $M$ means more computation time) as the results with higher $M$ were completely equal. It is also possible to set $M$ to a smaller value but this adds some noise leading to slightly poorer results.


\section{Evaluation} \label{eval}
In this section, we evaluate the probabilistic active learning for active class selection (PAL-ACS)-approach against other methods on multiple datasets in experimental comparisons. After describing our evaluation setup, we provide learning curves as well as error and sampling proportion tables and discuss the results.

\subsection{Evaluation Setup} \label{eval:settings}
\begin{figure}[h]
  \centering
  \subfloat[3Clusters]{
    \includegraphics[width=.3\linewidth]{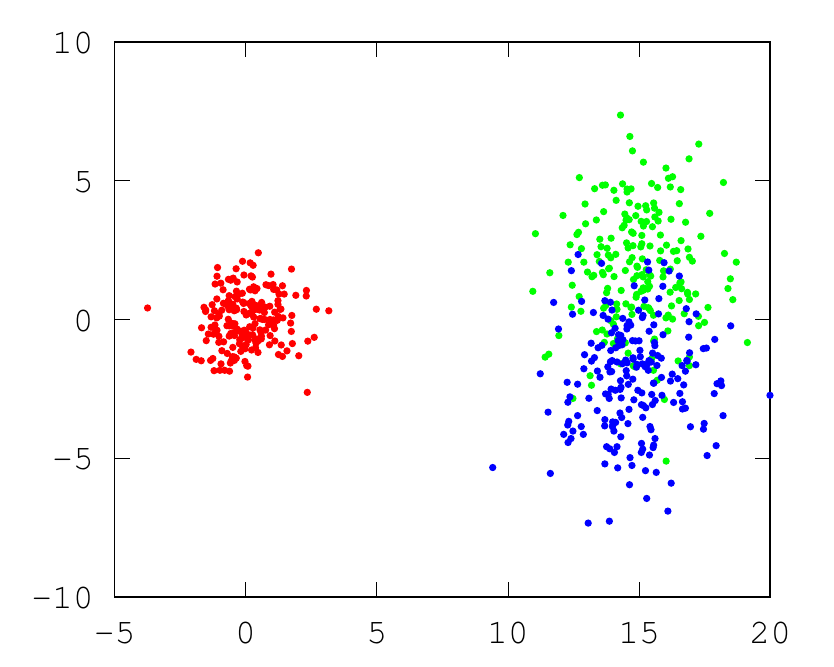}
    \label{fig:3clusters}
  }
  \subfloat[Spirals]{
    \includegraphics[width=.3\linewidth]{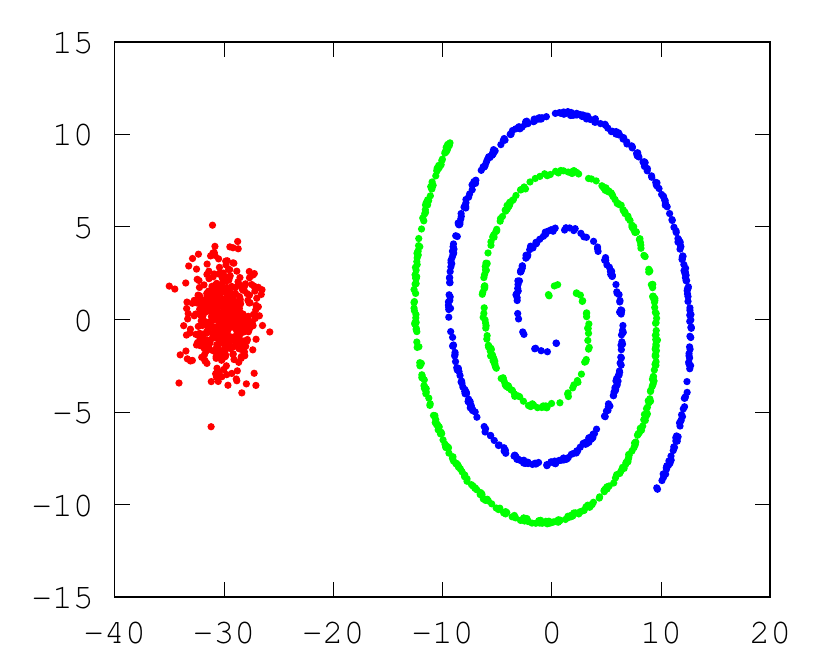}
    \label{fig:spirals}
  }
  \subfloat[Bars]{
    \includegraphics[width=.3\linewidth]{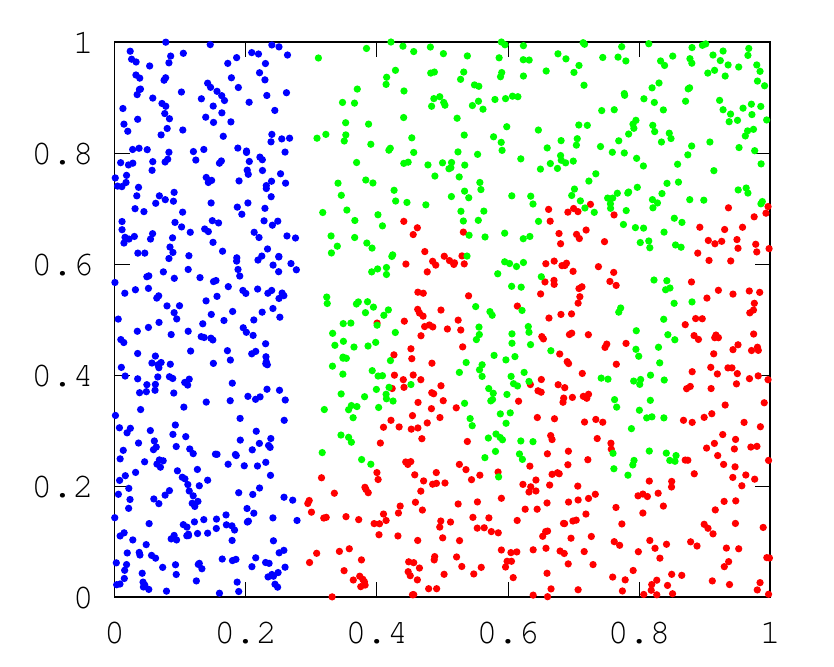}
    \label{fig:bars}
  }
  \caption{Scatterplots of the synthetic datasets.}
\end{figure}

The methods are evaluated on six different datasets. Thereof, three datasets are synthetic, having one class that is easily distinguishable from the others and two classes with a more complex decision boundary. A visualization of these two-dimensional datasets, called 3Clusters, Spirals and Bars, is given in Fig.~\ref{fig:3clusters}$\textendash$\ref{fig:bars}.
Additionally, we used three real-world datasets from the UCI machine learning repository \cite{UCI}, namely Vehicle, Vertebral Column, and Yeast. 
For Yeast, we selected five classes for our application: CYT, NUC, ME1, ME2, and ME3. 
We set the maximum number of learning steps, i.e. the budget depending on the complexity of the datasets to:  60 for 3Clusters, Vertebral, and Yeast, 80 for Vehicle, and 120 for Bars and Spirals.

As a baseline approach, we implemented the selecting strategy \emph{Random} that requests each class with equal probability. Furthermore, we compare against the state-or-the-art approaches \emph{Inverse} and \emph{Redistricting} published in \cite{LomaskyBrodleyAerneckeEtAl2007}. 

In a pre-processing step, all features are normalized to a $[0,1]$ range. For each dataset, we generated 500 random test-training-set combinations (trials). A test set consisting of 50 instances per class is extracted from the data, all remaining instances are used for training. To classify unseen data from the test set, we use a Parzen window classifier \cite{Chapelle2005,Parzen1962} with the same kernel used in the kernel frequency estimation. Due to feature normalization, the use of a constant kernel width for all datasets is reasonable, which we set to $\sigma = 0.05$. Error rate is used as performance measure and averaged over the 500 trials.

To ensure that only an algorithm's sampled class distribution influences its classification performance, we decided to use a fixed order of training instances per trial. When an algorithm requests an instance, the first instance of this class is returned. As a consequence, the training data obtained by different algorithms might overlap largely. Consider an example, where one ACS algorithm samples equally while the other samples 40\% from class 1 and 2 and 20\% from class 3. Although their sampled class distributions differ considerably, 26 of their first 30 acquired instances are completely equal. The fact that the resulting classifiers are therefore similar should be considered when reading the evaluation in the next chapter.

\subsection{Results and Discussion} \label{eval:results}

\begin{figure}[h]
  \centering
  \includegraphics[width=.32\linewidth,trim=1.6cm 6.5cm 1.6cm 6.3cm,clip]{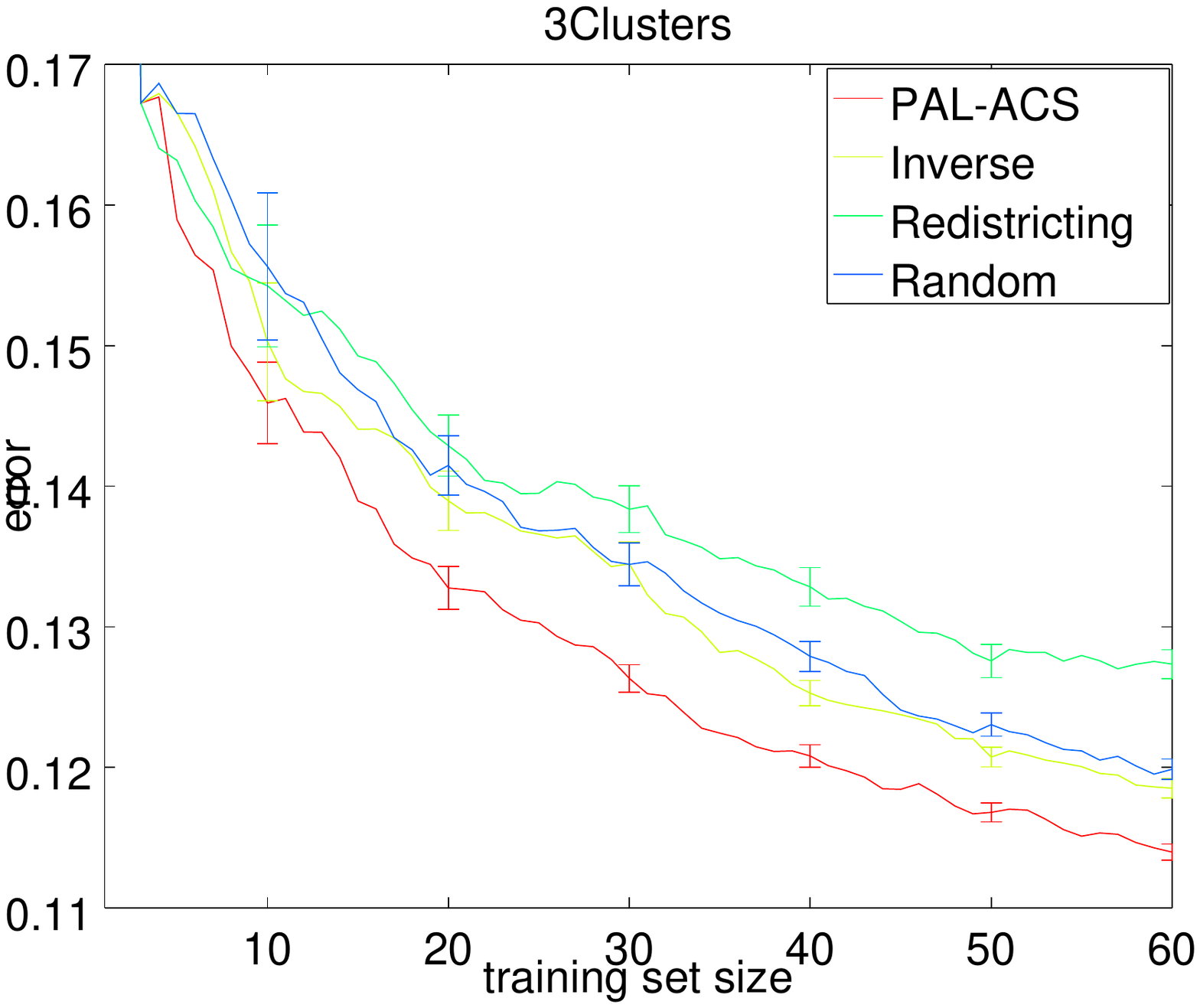}
  \includegraphics[width=.32\linewidth,trim=1.6cm 6.5cm 1.6cm 6.3cm,clip]{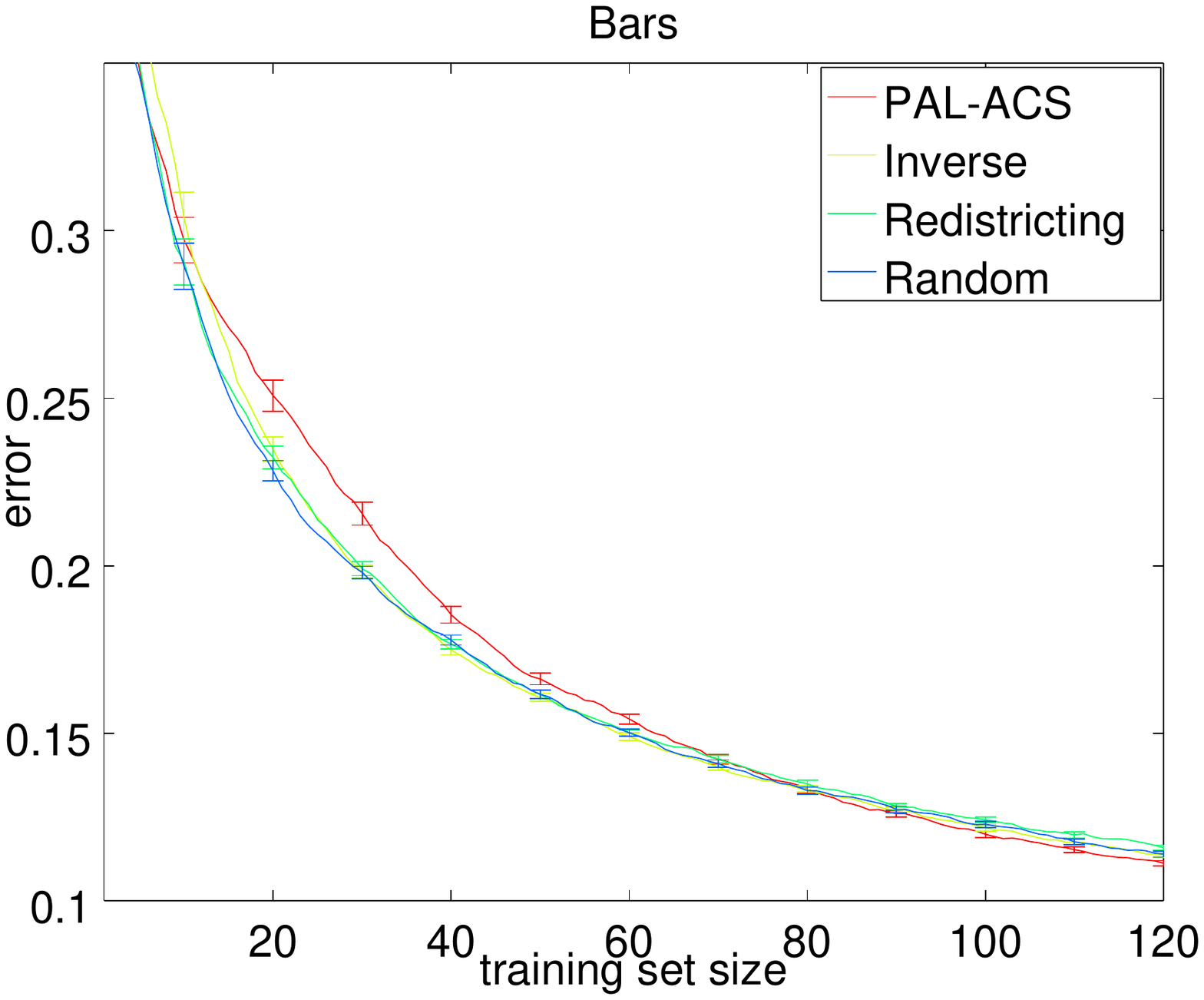}
  \includegraphics[width=.32\linewidth,trim=1.6cm 6.5cm 1.6cm 6.3cm,clip]{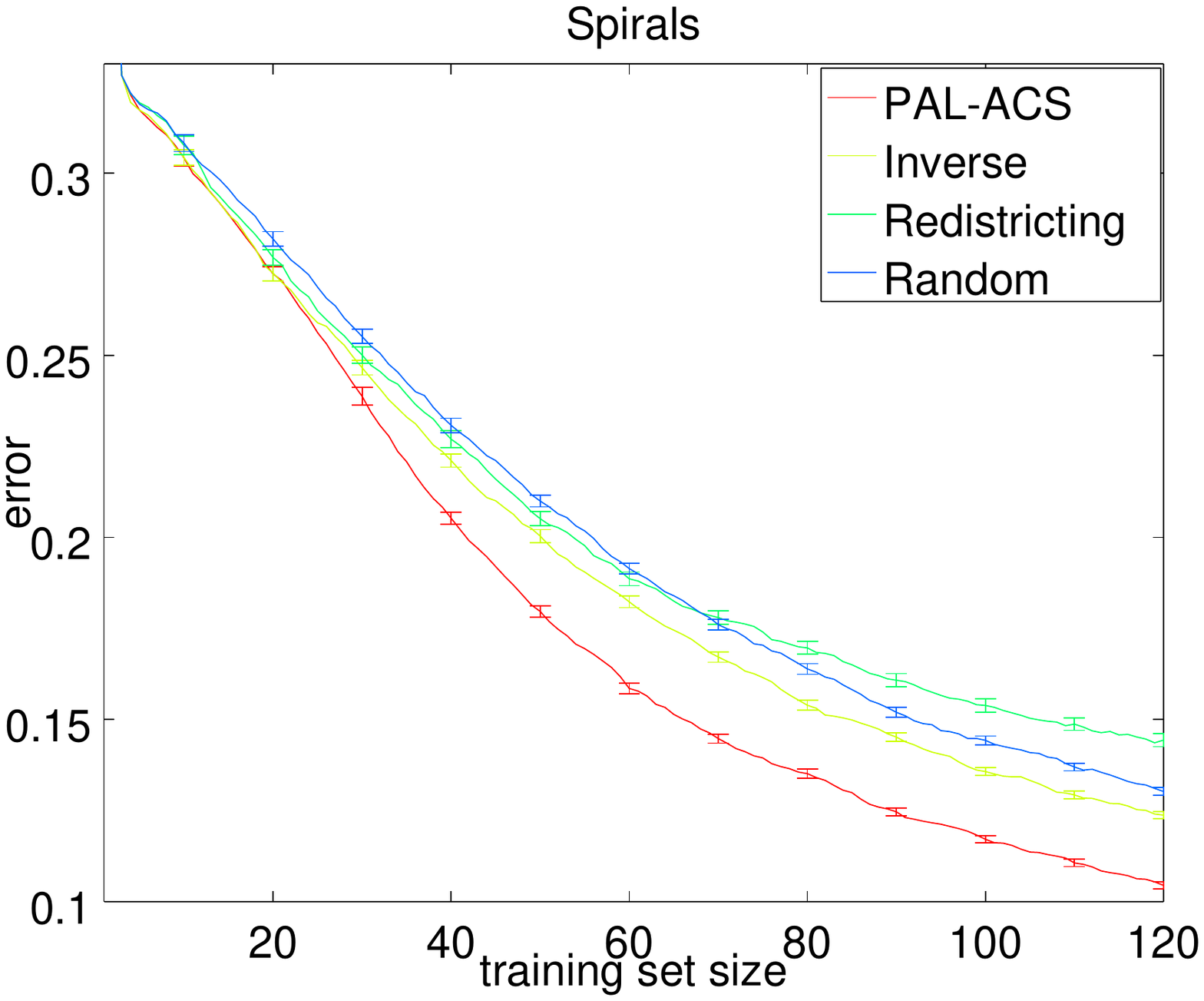}\\
  \includegraphics[width=.32\linewidth,trim=1.6cm 6.5cm 1.6cm 6.3cm,clip]{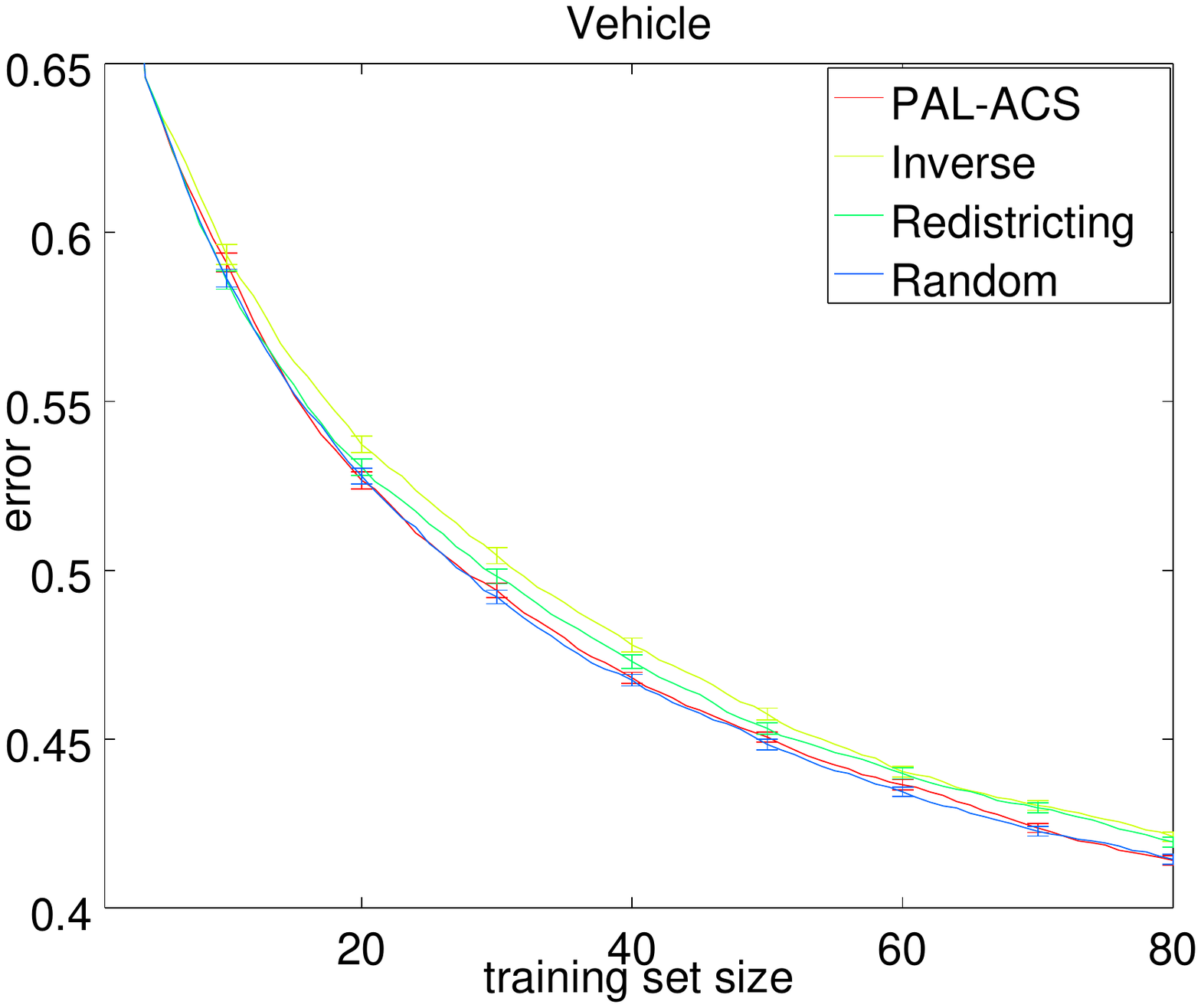}
  \includegraphics[width=.32\linewidth,trim=1.6cm 6.5cm 1.6cm 6.3cm,clip]{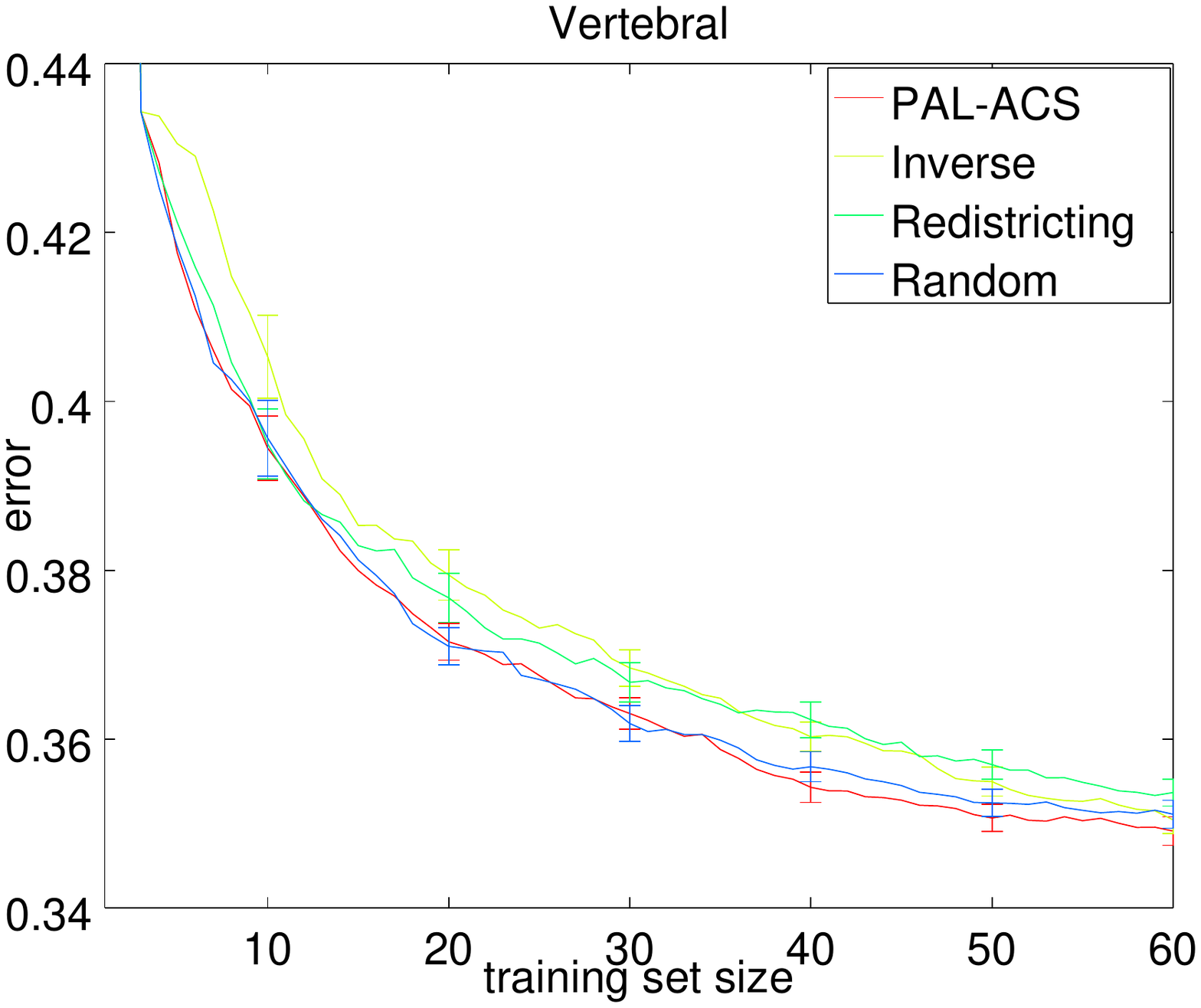}
  \includegraphics[width=.32\linewidth,trim=1.6cm 6.5cm 1.6cm 6.3cm,clip]{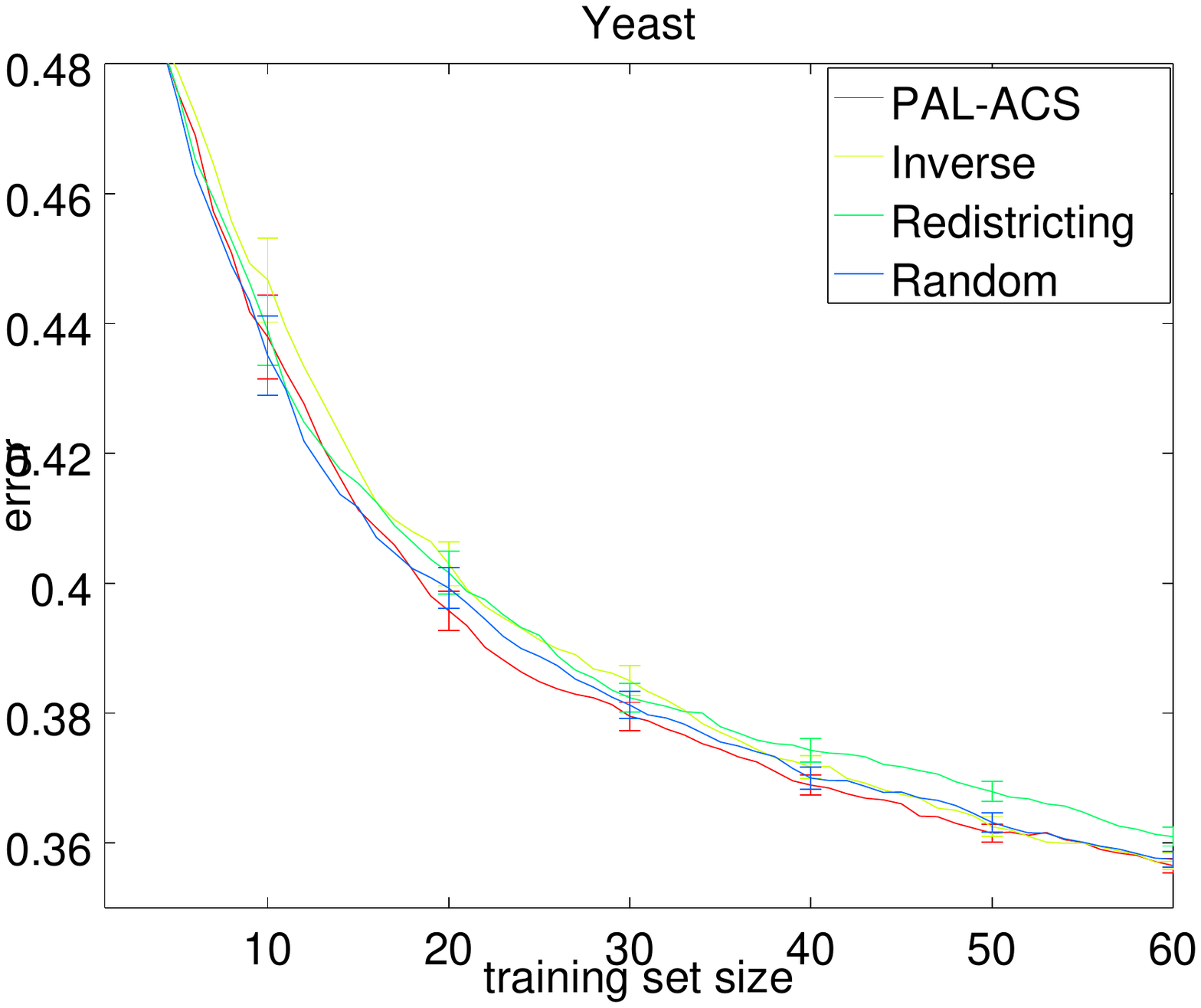}
  \caption{Learning Curves for each algorithm on every dataset. Each curve shows the mean error and standard deviation.}
  \label{fig:learningcurves}
\end{figure}

To compare the algorithms, we provide learning curves in Fig. \ref{fig:learningcurves}. These learning curves show the mean error and the variance of all algorithms with respect to the number of acquired instances. The best algorithm is the one that converges fastest to the lowest error. 

\begin{table}[t]
  \scriptsize
  \centering
\begin{tabular}{l|l|l|l|l|l|l|l|l|l}
Dataset & Method        & \multicolumn{2}{|c}{phase 1} & \multicolumn{2}{|c}{phase 2} & \multicolumn{2}{|c}{phase 3} & \multicolumn{2}{|c}{phase 4}\\
        &               & error & win ratio & error & win ratio & error & win ratio & error & win ratio \\ \hline
\hline
3Clusters & PAL-ACS       & $\mathbf{0.1498}$ & $\mathbf{40.97\%}$  & $\mathbf{0.1316}$ & $\mathbf{42.69\%}$  & $\mathbf{0.1215}$ & $\mathbf{45.85\%}$  & $\mathbf{0.1161}$ & $\mathbf{47.93\%}$  \\ 
          & Inverse       & $0.1543$ & $38.97\%$  & $0.1382$ & $33.67\%$  & $0.1271$ & $34.28\%$  & $0.1206$ & $34.17\%$  \\ 
          & Redistricting & $0.1557$ & $36.10\%$  & $0.1418$ & $30.25\%$  & $0.1339$ & $31.28\%$  & $0.1281$ & $31.69\%$  \\ 
          & Random        & $0.1575$ & $35.18\%$  & $0.1390$ & $31.13\%$  & $0.1294$ & $30.93\%$  & $0.1217$ & $32.68\%$  \\ 
\hline
Bars & PAL-ACS       & $0.2705$ & $25.97\%$  & $0.1773$ & $28.82\%$  & $0.1378$ & $32.47\%$  & $\mathbf{0.1177}$ & $\mathbf{38.22\%}$  \\ 
     & Inverse       & $0.2636$ & $31.12\%$  & $\mathbf{0.1686}$ & $33.14\%$  & $\mathbf{0.1364}$ & $32.24\%$  & $0.1196$ & $31.85\%$  \\ 
     & Redistricting & $0.2564$ & $34.73\%$  & $0.1697$ & $\mathbf{33.69\%}$  & $0.1384$ & $32.74\%$  & $0.1218$ & $30.67\%$  \\ 
     & Random        & $\mathbf{0.2539}$ & $\mathbf{35.59\%}$  & $0.1694$ & $32.65\%$  & $0.1371$ & $\mathbf{32.88\%}$  & $0.1202$ & $31.61\%$  \\ 
\hline
Spirals & PAL-ACS       & $\mathbf{0.2816}$ & $\mathbf{38.82\%}$  & $\mathbf{0.1927}$ & $\mathbf{58.07\%}$  & $\mathbf{0.1397}$ & $\mathbf{66.61\%}$  & $\mathbf{0.1139}$ & $\mathbf{66.81\%}$  \\ 
        & Inverse       & $0.2831$ & $34.47\%$  & $0.2103$ & $24.53\%$  & $0.1611$ & $21.94\%$  & $0.1328$ & $20.74\%$  \\ 
        & Redistricting & $0.2861$ & $30.61\%$  & $0.2165$ & $21.17\%$  & $0.1735$ & $15.96\%$  & $0.151$ & $15.64\%$  \\ 
        & Random        & $0.2897$ & $26.16\%$  & $0.2205$ & $13.66\%$  & $0.1701$ & $13.09\%$  & $0.1404$ & $13.97\%$  \\ 
\hline
Vehicle & PAL-ACS       & $0.5783$ & $\mathbf{34.45\%}$  & $0.4931$ & $29.19\%$  & $0.4499$ & $29.89\%$  & $0.4238$ & $32.06\%$  \\ 
        & Inverse       & $0.5851$ & $31.49\%$  & $0.5041$ & $23.60\%$  & $0.4572$ & $23.04\%$  & $0.4301$ & $23.75\%$  \\ 
        & Redistricting & $0.5783$ & $33.62\%$  & $0.4981$ & $26.00\%$  & $0.4536$ & $28.84\%$  & $0.4290$ & $26.78\%$  \\ 
        & Random        & $\mathbf{0.5776}$ & $34.32\%$  & $\mathbf{0.4920}$ & $\mathbf{33.42\%}$  & $\mathbf{0.4486}$ & $\mathbf{32.75\%}$  & $\mathbf{0.4230}$ & $\mathbf{33.89\%}$  \\ 
\hline
Vertebral & PAL-ACS       & $\mathbf{0.3989}$ & $34.00\%$  & $0.3696$ & $31.55\%$  & $\mathbf{0.3566}$ & $\mathbf{33.44\%}$  & $\mathbf{0.3506}$ & $\mathbf{33.63\%}$  \\ 
          & Inverse       & $0.4088$ & $30.53\%$  & $0.3764$ & $27.61\%$  & $0.3625$ & $27.97\%$  & $0.3536$ & $28.99\%$  \\ 
          & Redistricting & $0.4009$ & $34.13\%$  & $0.3737$ & $30.12\%$  & $0.363$ & $28.57\%$  & $0.3557$ & $27.25\%$  \\ 
          & Random        & $0.3993$ & $\mathbf{34.38\%}$  & $\mathbf{0.3695}$ & $\mathbf{32.43\%}$  & $0.3578$ & $32.51\%$  & $0.3522$ & $31.85\%$  \\ 
\hline
Yeast & PAL-ACS       & $0.4439$ & $37.38\%$  & $\mathbf{0.3909}$ & $\mathbf{29.99\%}$  & $\mathbf{0.3716}$ & $\mathbf{29.59\%}$  & $\mathbf{0.3606}$ & $31.20\%$  \\ 
      & Inverse       & $0.4495$ & $32.48\%$  & $0.3967$ & $27.23\%$  & $0.3744$ & $28.80\%$  & $0.3612$ & $\mathbf{31.64\%}$  \\ 
      & Redistricting & $0.4444$ & $35.07\%$  & $0.3958$ & $26.99\%$  & $0.3762$ & $27.39\%$  & $0.3659$ & $23.67\%$  \\ 
      & Random        & $\mathbf{0.4417}$ & $\mathbf{38.40\%}$  & $0.3931$ & $28.48\%$  & $0.3731$ & $27.95\%$  & $0.3617$ & $28.21\%$  \smallskip \\ 
\end{tabular}
  \medskip
  \caption{Quantitative Comparison of ACS methods on all datasets. To show the learning process, the mean of errors has been calculated for different learning phases. Additionally, the ratio of won trials for each algorithm is shown. The winner is printed in bold.}
  \label{tab:acc}
\end{table}

Additionally, we provide the quantitative values for the algorithms’ performances in Tab. \ref{tab:acc}. Here, we separated the learning process into four phases, in order to determine how fast algorithms get the structure of the learning problem. Each phase contains $25\%$ of the learning steps. For each phase, we determine the mean accuracy for each algorithm on each dataset and calculate the ratio of won trials. Note, that these ratios do not sum to one because some trials have multiple winners due to the aspects discussed at the end of Sec.~\ref{eval:settings}.

%
%

First of all, the result from the plots and the table show that the current state-of-the-art approaches do not achieve considerable better results than random sampling, which justifies the development of our new ACS method.   
PAL-ACS is constantly better than both competing ACS methods with one exception. In the Bars dataset, our method performed better only towards the very end. This might be due to the non-Gaussian structure of the data as PAL-ACS internally uses Gaussian kernels to generate the pseudo instances. Comparing PAL-ACS to Random, we see that the superiority of our method depends on the structure of the data. The higher the differences in the complexity of the classes, the more beneficial is PAL-ACS.

PAL-ACS’s high performance can be explained when looking at the sampled class distributions in Tab. \ref{tab:samplingprop}. 
The results on 3Clusters, Spirals, and Bars in Tab.~\ref{tab:samplingprop} show that PAL-ACS contributes a smaller
sampling proportion to the easier class (1\textsuperscript{st} in 3Clusters and Spirals, 3\textsuperscript{rd} in Bars) than to the more
difficult ones. Inverse and Redistricting show the same tendency, but to a much smaller extend, resulting in weaker performance in 3Clusters and Spirals. Although PAL-ACS showed mediocre results in Bars, it determined the easy class even in early phases.

\begin{table}[htb]
\small
\begin{tabular}{l|c|c|c|c|c|c}
Method      & 3Clusters & Bars & Spirals &  Vehicle & Vertebral & Yeast \\ \hline
PAL-ACS       ~&~ 17,42,41 ~&~ 38,41,21 ~&~ 05,49,46 ~&~ 25,25,25,25 ~&~ 30,35,34 ~&~ 23,27,27,23 \\
Inverse       ~&~ 29,35,36 ~&~ 35,36,30 ~&~ 28,36,36 ~&~ 27,27,24,23 ~&~ 38,36,25 ~&~ 28,28,22,23 \\
Redist.       ~&~ 25,37,38 ~&~ 38,37,24 ~&~ 19,41,39 ~&~ 26,26,23,25 ~&~ 38,37,25 ~&~ 29,27,20,24 \\
Random        ~&~ 33,34,33 ~&~ 33,33,34 ~&~ 33,34,33 ~&~ 25,25,25,25 ~&~ 33,34,33 ~&~ 25,25,25,25 
\end{tabular}
\smallskip
\captionof{table}{Final sampling proportions (for all classes) in percent.}
\label{tab:samplingprop}
\end{table}

Vehicle as a dataset with equally difficult classes demonstrates PAL-ACS's ability to detect this fact and converge to a uniform sampling proportion. Random performs slightly better on the Vehicle dataset as it has the advantage of assuming classes to be equally difficult per default, Inverse and Redistricting yield worse results by undersampling classes.
On Vertebral and Yeast, we can see a clear sampling tendency in Tab.~\ref{tab:samplingprop} which is also visible in the results of the learning curves in Fig.~\ref{fig:learningcurves}, resp. the performance table (Tab.~\ref{tab:acc}).

Overall, PAL-ACS always identifies the difficult classes and samples accordingly. As a result, its
performance is best (in cases some classes are more difficult than others) or equal with the best
competitor Random (in cases all classes are equally difficult).

\section{Conclusion} \label{conc}
In this paper, we introduced a new approach for active class selection, called PAL-ACS (probabilistic active learning for active class selection). This method is based on the performance gain function proposed in \cite{KottkeKrempl2016ECAI} which was originally introduced for active learning. To apply this function, the ACS problem has been transformed in an active learning problem by generating pseudo instances. 

The experimental evaluation shows our method's superiority on datasets where a non-uniform sampling improves the classifier's performance. On datasets with equally complex classes, our method identifies uniform sampling to be the best. Thus, in contrast to other active class selection methods, it performs comparably well with random sampling which is a uniform sampler per default.

In the future, we want to combine this approach with more sophisticated learning models and evaluate our algorithm on further datasets and real world BCI data. An interesting topic is the comparison of our usefulness model with human information acquisition as mentioned in \cite{MarkantSettlesGureckis2016}.

\subsection*{Acknowledgements}
We thank the Psychoinformatics lab, esp. Michael Hanke and Alex Waite, from Magdeburg University to let us use their cluster, our colleague Pawel Matuszyk, and the reviewers for their inspiring comments.


\newpage \clearpage
\bibliographystyle{plain}
\bibliography{palacs}

\end{document}